# Slope stability predictions on spatially variable random fields using machine learning surrogate models

Mohammad Aminpour*, PhD, Research Assistant (Corresponding author)

Civil and Infrastructure Engineering Discipline, School of Engineering, Royal Melbourne Institute of Technology (RMIT), Victoria 3001, Australia

E-mail: mohammad.aminpour@rmit.edu.au

Reza Alaie, PhD Graduate

Department of Civil Engineering, Faculty of Engineering, University of Guilan, Rasht, Iran

E-mail: alaiereza@phd.guilan.ac.ir

Navid Kardani, PhD Research Fellow

Civil and Infrastructure Engineering Discipline, School of Engineering, Royal Melbourne Institute of Technology (RMIT), Victoria 3001, Australia

Email: navid.kardani@rmit.edu.au

Sara Moridpour, PhD, Associate Professor

Civil and Infrastructure Engineering Discipline, School of Engineering, Royal Melbourne Institute of Technology (RMIT), Victoria 3001, Australia

Email: sara.moridpour@rmit.edu.au

Majidreza Nazem, PhD, Associate Professor

Civil and Infrastructure Engineering Discipline, School of Engineering, Royal Melbourne Institute of Technology (RMIT), Victoria 3001, Australia

Email: Majidreza.nazem@rmit.edu.au

# Slope stability predictions on spatially variable random fields using machine learning surrogate models

Mohammad Aminpour[1*], Reza Alaie[2], Navid Kardani[1], Sara Moridpour[1], Majidreza Nazem[1]

[1] Civil and Infrastructure Engineering, School of Engineering, RMIT University, Melbourne, Australia

[2] Department of Civil Engineering, Faculty of Engineering, University of Guilan, Rasht, Iran.

*Corresponding author:* Mohammad Aminpour, E-mail: mohammad.aminpour@rmit.edu.au

**Abstract:**

Random field Monte Carlo (MC) reliability analysis is a robust stochastic method to determine the probability of failure. This method, however, requires a large number of numerical simulations demanding high computational costs. This paper explores the efficiency of different machine learning (ML) algorithms used as surrogate models trained on a limited number of random field slope stability simulations in predicting the results of large datasets. The MC data in this paper require only the examination of failure or non-failure, circumventing the time-consuming calculation of factors of safety. An extensive dataset is generated, consisting of 120,000 finite difference MC slope stability simulations incorporating different levels of soil heterogeneity and anisotropy. The Bagging Ensemble, Random Forest and Support Vector classifiers are found to be the superior models for this problem amongst 9 different models and ensemble classifiers. Trained only on 0.47% of data (500 samples), the ML model can classify the entire 120,000 samples with an accuracy of %85 and AUC score of %91. The performance of ML methods in classifying the random field slope stability results generally reduces with higher anisotropy and heterogeneity of soil. The ML assisted MC reliability analysis proves a robust stochastic method where errors in the predicted probability of failure using %5 of MC data is only %0.46 in average. The approach reduced the computational time from 306 days to less than 6 hours.

**Highlights**

- Nine machine Learning (ML) classifiers are developed as surrogate models for slope stability reliability analysis.
- To validate the ML models, a complete Monte Carlo (MC) dataset containing 120,000 anisotropic heterogeneous slopes is generated.
- Compared to complete MC datasets, surrogate models prove a robust stochastic reliability method with high accuracies.
- Using 5% of MC data, the error of ML predicted probability of failure for all anisotropy and heterogeneity levels is below 1% with an average of 0.46%.
- With no need to calculate the factors of safety, the approach reduced the computational time from 306 days to less than 6 hours.

# 1. Introduction

The uncertainty of geotechnical parameters is the core of reliability-based design in geotechnics. The uncertainty can be associated with inherent spatial variability of soil or deterministic calculation models in transforming measurement information into design parameters (Phoon and Kulhawy 1999, Juang, Zhang et al. 2019). The inherent variability of geotechnical parameters is known as the major source of uncertainty (Fenton and Griffiths 2008) with a significant influence on slope stability (Christian, Ladd et al. 1994, Ching and Phoon 2013, Lloret-Cabot, Fenton et al. 2014), Lloret-Cabot, Fenton et al. 2014). However, while being critically important in a safe geotechnical design, reliability analysis can be practically difficult to implement due to the great computational efforts required. Combined with the random field theory (Vanmarcke 1977, Vanmarcke 1977) and the finite element method, the Monte Carlo (MC) approach as a conceptually simple and robust statistical method (Wang, Cao et al. 2010, Wang, Cao et al. 2011) involves the production of thousands of simulations to evaluate the probability of geotechnical events such as slope failure. In small probability events which are of high importance in geotechnical practice, the MC method can be computationally inefficient. The significant computational time and power needed for this method enforce geotechnical engineers to trade off the risk management for timely progress of project design. However, underestimating the impacts of geotechnical uncertainties can remarkably increase the associated risks with examples in tunnelling (Chen, Wang et al. 2019), slope stability (Cho 2007, Wang, Hwang et al. 2013), Wang, Hwang et al. 2013, Gong, Tang et al. 2020), bearing capacity (Li, Tian et al. 2015), etc.

Different techniques are introduced to address the computational challenges of geotechnical MC reliability assessments, including the variance reduction methods such as importance sampling (Au and Beck 2003) or subset simulation (Au and Beck 2001, Papaioannou, Betz et al. 2015), Papaioannou, Betz et al. 2015). It is shown that the importance sampling is not applicable in problems with a large number of random variables (Huang, Fenton et al. 2017). Also, the subset simulation which has been applied for small failure probability problems in slope stability (Li, Xiao et al. 2016, Tian, Li et al. 2021), appeared to be even more time consuming than an MC simulation for a given accuracy when a search for the factor of safety (FOS) is required (Huang, Fenton et al. 2017). The subset simulation is enhanced by avoiding a search for FOS using the strength reduction technique to increase the efficiency of this method. However, even with such enhancement, the subset simulation has not reduced the simulation time more than 3 times and can be still less efficient than a complete MC simulation in many cases (Huang, Fenton et al. 2017).

Another technique to increase the efficiency of reliability analysis of slope stability is the construction of surrogate models or response surface methods, including polynomial chaotic expansion, support vector regression, Kriging model, and multiple adaptive regression spline (Jiang, Li et al. 2014, Jiang and Huang 2016, Kang, Xu et al. 2016, Liu and Cheng 2018, Liu, Zhang et al. 2019, Wang, Wu et al. 2020, Zeng, Zhang et al. 2020, Deng, Pan et al. 2021). This technique involves the introduction of surrogate models to replace the Finite Element simulations approximating the relationship between the input variables and the output response. Among the surrogate models, machine learning algorithms have received increasing attention in slope stability problems with spatial variability (Kang, Xu et al. 2016, Zhu, Pei et al. 2019, He, Xu et al. 2020, He, Wang et al. 2021, He, Wang et al. 2021, Wang and Goh 2021, Zhang, Phoon et al. 2021, Zhu, Hiraishi et al. 2021). ML methods have been widely used in slope stability predictions and landslide characterisations (Wei, Lü et al. 2019, Deng, Smith et al. 2021, Han, Shi et al. 2021, Hu, Wu et al. 2021, Huang, Han et al. 2021, Sun, Xu et al. 2021, Wang, Zhang et al. 2021, Zhao, Meng et al. 2021). However, a systematic study on different ML models when used as surrogate models and in particular, the dependence of their performance on various levels of heterogeneity and anisotropy of random fields is yet to be performed. Moreover, previous studies have normally trained the ML models on the data of Monte Carlo simulations with calculated FOSs which are tedious tasks in geotechnical simulations. However, in a direct MC simulation, the probability of failure can be examined without a need for the calculation of FOS in each simulation (Huang, Fenton et al. 2017). Conducted either by numerous failure surfaces in a limit equilibrium analysis or using the strength reduction method, the calculation of FOS can significantly increase the computational cost of the MC method. Little is known about the performance of ML surrogate models on slope stability MC data without determined FOSs.

In this study, we systematically investigate the performance of different ML surrogate models on spatially variable random field MC data with no FOS calculated. The selected ML algorithms include six models and three ensemble classifiers which have been used in geotechnical engineering for more than a decade with their appropriateness being shown in slope stability studies (Das, Sahoo et al. 2010, Samui and Kothari 2011, , Samui and Kothari 2011, Cheng and Hoang 2016, Pham, Bui et al. 2017, Feng, Li et al. 2018, , Pham, Bui et al. 2017, Feng, Li et al. 2018, Bui, Nguyen et al. 2020, Kardani, Zhou et al. 2021, Zhang, Wu et al. 2021), Kardani, Zhou et al. 2021, Zhang, Wu et al. 2021). These models are trained on small fractions of MC data when the outcome of MC simulations are only the failure or non-failure status of slopes, thus further increasing the efficiency with the less computational time needed for calculated FOSs. The performance of models is tested against a complete MC database covering a wide range of random field heterogeneity and anisotropy. The sensitivity of results is assessed using the repeated k-fold cross-validation technique.

This paper is structured as follows. Section 2 describes the technical method to generate the random fields. Section 3 introduces the random field finite difference computational method to generate slope models. The details of the generated Monte Carlo data with statistical parameters are summarised in Section 4. An overview on the evaluation approach of ML surrogate models implemented in this study is given in Section 5. The machine learning algorithms and ensemble classifiers are briefly described in Section 6. Section 7 provides the results comparing the performance of different ML classifiers when used as surrogate models on various datasets with respect to the levels of anisotropy and heterogeneity using suitable performance scores. The sensitivity of the results is also discussed. A summary with an overview of the future research directions is presented in Section 8.

## 2. Random field generation

A random field as a random function defined over an arbitrary domain is used to model the spatial variability of geotechnical parameters. Statistical parameters including mean value µ, standard deviation σ and the correlation distance δ are used to characterise the random fields. The correlation distance, also known as the scale of fluctuation, characterises the variation of the field in space which is captured by the covariance function. Assuming a Gaussian and stationary random field where the complete probability distribution is independent of absolute location, the undrained shear strength parameter, $C_u$, is the random variable in this study.

Among several methods for the generation of random fields such as Covariance Matrix Decomposition, Moving Average, Fast Fourier Transform, Turning Bands Method and Local Average Subdivision Method (Griffiths and Fenton 2007), the Cholesky decomposition technique is selected and used here which is developed in Ref. (Vetterling, Press et al. 2002) and utilised in various geotechnical studies (Zhu and Zhang 2013) (Zhu, Zhang et al. 2017).

In this paper, the random field is defined as (El-Kadi and Williams 2000)

$$C_u(\tilde{x}) = exp\left(L.\varepsilon + \mu_{lnC_u(\tilde{x})}\right) \qquad (1)$$

where $C_u(\tilde{x})$ is the undrained shear strength of soil at the spatial position $\tilde{x}$, $\varepsilon$ is an independent random variable normally distributed with zero mean and unit variance, $\mu_{lnC_u(\tilde{x})}$ is the mean of the logarithm of $C_u$, and L is a lower-triangular matrix computed from the decomposition of the covariance matrix using the Cholesky decomposition technique (Griffiths and Fenton). The anisotropic exponential Markovian covariance function is defined as (Fenton and Griffiths 2008)

$$A(l_x, l_y) = \sigma_{\ln C_u}^2 \exp\left(-\frac{|l_x|}{\delta_h} - \frac{|l_y|}{\delta_v}\right) \tag{2}$$

where $l_x$ and $l_y$ are the horizontal and vertical distances between two arbitrary points, $\delta_h$ and $\delta_v$ are the horizontal and vertical scales of fluctuation (correlation distances), respectively, and $\sigma_{\ln C_u}$ is the standard deviation of the logarithms of undrained shear strength.

According to the Cholesky decomposition technique, the covariance between the logarithms of the random variable values (here $C_u$) at any two points is decomposed into

$$A = LL^T \tag{3}$$

where $T$ stands for transpose.

The mean ($\mu$) and the standard deviation ($\sigma$) of the logarithms of $C_u$ are computed as

$$\mu_{\ln C_u} = \ln \mu_{C_u} - \frac{1}{2}\sigma_{\ln C_u}^2 \tag{4}$$

$$\sigma_{\ln C_u} = \sqrt{\ln(1 + COV_{C_u}^2)} \tag{5}$$

where $COV_{C_u}$ is the coefficient of variation of $C_u$ which is obtained by

$$COV_{C_u} = \frac{\sigma_{C_u}}{\mu_{C_u}}. \tag{6}$$

Random filed simulations include a range of mean undrained shear strength values, $\mu_{C_u}$ from 18.6 to 33.5 kPa. These $\mu_{C_u}$ values are chosen based on the FOSs associated with homogeneous slopes where the shear strength of soil is equal to $\mu_{C_u}$. The FOS for a homogeneous slope with $C_u = 18.6$ kPa is equal to 1. The other $\mu_{C_u}$ values are determined as 1.2, 1.4, 1.6 and 1.8 times of 18.6 kPa. The COV values are changed as 0.1, 0.3 and 0.5 representing low to high levels of heterogeneity. The vertical correlation distance, $\delta_v$ is kept constant as 1 m, whereas the horizontal correlation distance, $\delta_h$ varies from 1 to 25 m. Thus, the anisotropy ratio, $\xi = \frac{\delta_h}{\delta_v}$ varies from 1 to 25. The summary of statistical parameters used in this study is presented in Table .

Table 1. The statistical parameters of spatial variability of soil in this study

| Parameter | Values |
| --- | --- |
| Mean of undrained shear strength, $\mu_{C_u}$ (kPa) | 18.6, 22.3, 26, 29.7, 33.5 |
| Coefficient of variation, $COV_{C_u}$ | 0.1, 0.3, 0.5 |
| Horizontal correlation distance, $\delta_h$, m | 1, 6, 12, 25 |
| Vertical correlation distance, $\delta_v$, m | 1 |

## 3. Finite difference random field model

The slope considered for this study, as shown in Fig. 1, has the following geometrical specifications: the slope angle of 45°, the slope height of 5 m and the foundation depth of 10 m from the top of the slope. The foundation level is assigned with a rigid boundary surface simulating the hard rock bed with all movements restricted. The horizontal displacements are also restricted at the vertical boundaries. A linear elastic-perfectly plastic stress-strain behaviour incorporating the Mohr-Coulomb failure criterion is adapted for the soil. The shear modulus of soil $G$ is assumed to be a linear function of the undrained shear strength as $G = I_R C_u$ where $I_R$ is the rigidity index. $I_R$ is defined as the ratio between the shear modulus G and undrained shear strength. Here, $I_R = 800$ denoting a moderately stiff soil is assumed according to the ranges suggested in the literature (e.g., $I_R = 300 - 1500$ (Popescu, Deodatis et al. 2005)). The soil unit weight equal to $\gamma_{sat} = 20 \text{kN/m}^3$ is kept constant in all simulations.

FLAC (FLAC Itasca) is used for random finite difference simulations. The mesh size was tested to be sufficiently fine in acquiring accurate results with less than 5% relative errors compared to a highly fine but computationally expensive mesh size. A four-nodded quadrilateral grid with a 0.5m side dimension was generated for this study resulting in 800 cells. Each cell attributes to a different $C_u$ value in a random field realisation (Fig. 1).

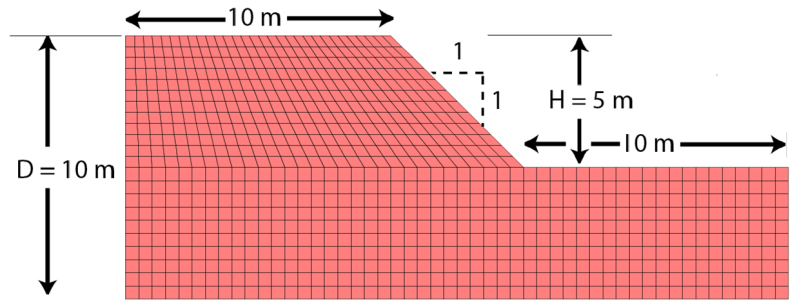

**Fig. 1** The slope random finite difference model comprising 800 random variable cells

To identify if a slope is stable or not, an alternative criterion for slope failure is adapted replacing the calculation of FOS using the traditional time-consuming strength reduction technique. In this method, a combined criterion for plastic zones and the velocity of grids is assessed in the finite difference model. The failure surface can be detected if a contiguous region of active plastic zones connecting two boundary surfaces is developed. For the velocity criterion, the failure is also determined when both the velocity gradient and amplitude are greater than the determined thresholds. The alternative failure criterion is shown to be efficient with accurate results as compared to stochastic stability results obtained using the strength reduction technique (see Ref. (Jamshidi Chenari and Alaie 2015) for further details).

## 4. Monte Carlo reliability analysis

Monte Carlo reliability analysis involves the simulation of a typically large number of samples for desired statistical parameters. The statistics of the results such as the probability of failure converges to a constant value in the large datasets. For this study, it can be shown that 2000 samples are sufficient (Jamshidi Chenari and Alaie 2015) for constant statistics of the results. This number agrees with the findings suggesting that a typical number of 2500 simulations can provide reasonable precision and reproducibility for random field finite element studies (Griffiths and Fenton 2007). Thus, for each set of statistical parameters $(\mu_{C_u}, COV_{C_u}, \delta_h, \delta_v)$, 2000 random field realisations are generated. Each realisation is simulated using the finite difference method to determine the stability status of slopes. The outcomes of all analyses as a binary result (failure or non-failure) provide the Monte Carlo data for the reliability analysis. A summary of the details of the random fields generated in this study is shown in Table 2.

**Table 2.** The characteristics of the random fields generated in this study

| Random field label | COV | $\delta_v$ (m) | $\xi$ | Mean of undrained shear strength, $\mu_{C_u}$ (kPa) | No. of random field simulations |
|---|---|---|---|---|---|
| $U_{0.1,1}$ | 0.1 | 1 | 1 | 18.6, 22.3, 26, 29.7, 33.5 | 5 × 2000 |
| $U_{0.1,6}$ | 0.1 | 1 | 6 | 18.6, 22.3, 26, 29.7, 33.5 | 5 × 2000 |
| $U_{0.1,12}$ | 0.1 | 1 | 12 | 18.6, 22.3, 26, 29.7, 33.5 | 5 × 2000 |
| $U_{0.1,25}$ | 0.1 | 1 | 25 | 18.6, 22.3, 26, 29.7, 33.5 | 5 × 2000 |
| $U_{0.3,1}$ | 0.3 | 1 | 1 | 18.6, 22.3, 26, 29.7, 33.5 | 5 × 2000 |
| $U_{0.3,6}$ | 0.3 | 1 | 6 | 18.6, 22.3, 26, 29.7, 33.5 | 5 × 2000 |
| $U_{0.3,12}$ | 0.3 | 1 | 12 | 18.6, 22.3, 26, 29.7, 33.5 | 5 × 2000 |
| $U_{0.3,25}$ | 0.3 | 1 | 25 | 18.6, 22.3, 26, 29.7, 33.5 | 5 × 2000 |
| $U_{0.5,1}$ | 0.5 | 1 | 1 | 18.6, 22.3, 26, 29.7, 33.5 | 5 × 2000 |
| $U_{0.5,6}$ | 0.5 | 1 | 6 | 18.6, 22.3, 26, 29.7, 33.5 | 5 × 2000 |
| $U_{0.5,12}$ | 0.5 | 1 | 12 | 18.6, 22.3, 26, 29.7, 33.5 | 5 × 2000 |
| $U_{0.5,25}$ | 0.5 | 1 | 25 | 18.6, 22.3, 26, 29.7, 33.5 | 5 × 2000 |
| | | | | total | 120000 |

**Error! Reference source not found.** illustrates some examples of random field realisations. Each plot in **Error! Reference source not found.** shows a single realisation of the random fields for the given $\mu_{C_u}$, COV and ξ parameters from 2000 realisations. The mean value of $C_u$ for all illustrated slopes is 18.6 kPa.

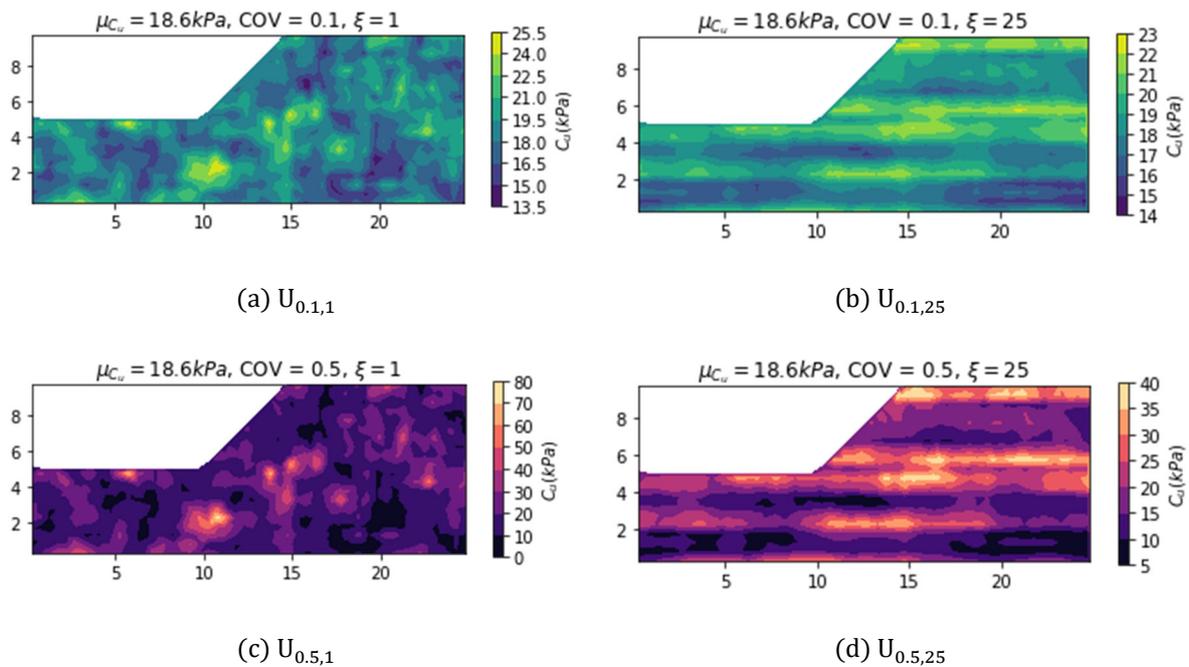

(a) $U_{0.1,1}$  (b) $U_{0.1,25}$

(c) $U_{0.5,1}$  (d) $U_{0.5,25}$

**Fig. 2** Examples of the random field slope models with spatial variability of the undrained shear strength of the soil.

## 5. Evaluation of ML surrogate models

In this paper, a complete Monte Carlo analysis on a typical slope stability problem in conducted. Covering a wide range of heterogeneity and anisotropy of random fields, the MC study provides a full-size database as a benchmark by which an actual evaluation of the trained ML surrogate models can be possible. The classification tasks and the probability of failure obtained from ML surrogate models using 5% of data are then compared with the actual results from the MC benchmark data. An overview of the implemented approach in this study is shown as a flowchart in Fig. 3.

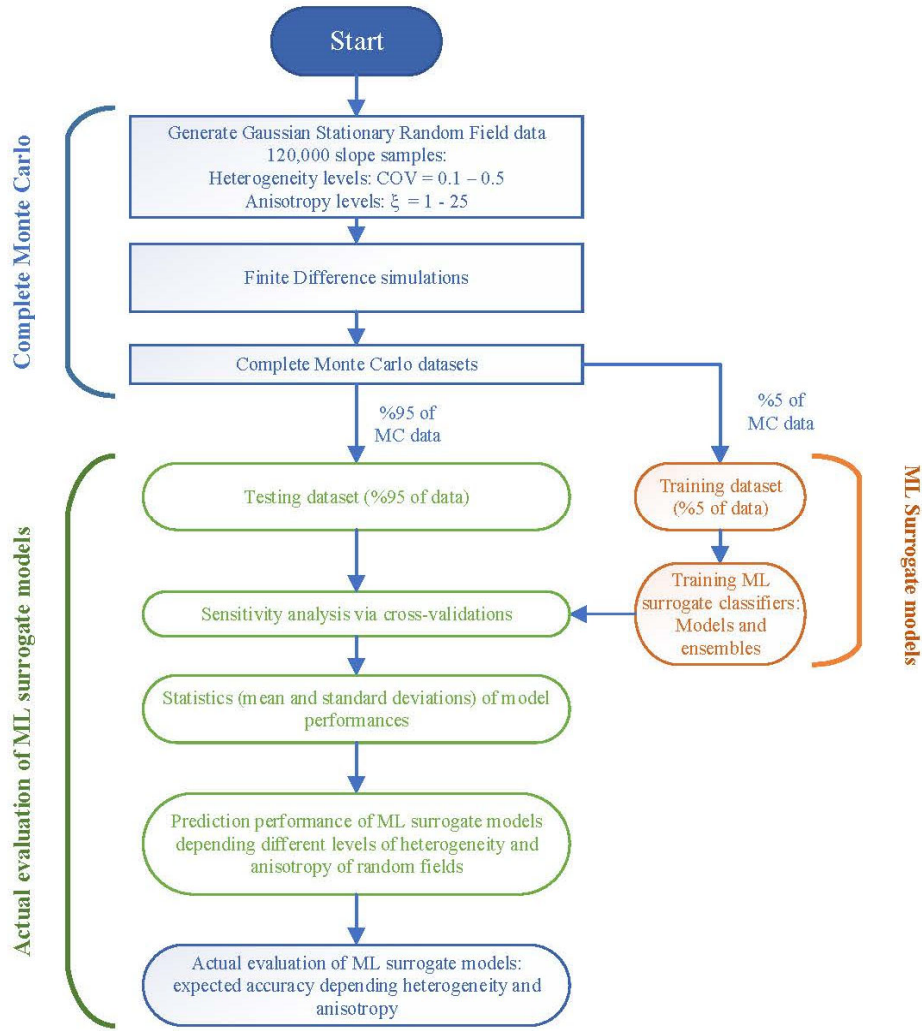

**Fig. 3** The approach implemented in this study in evaluating the ML surrogate models versus actual full-size MC datasets.

## 6. Machine Learning Algorithms

### 6.1. Logistic Regression

The possibility of a binary result can be predicted using the Logistic Regression (LR) algorithm. The LR is utilised to describe the connection between one dependent binary variable and one or more independent variables at the nominal, interval, ordinal, or ratio level. A binary logistic model mathematically consists of a dependent variable with two potential values indicated by an indicator variable with the values "0" and "1". A LR model generally estimates the likelihood of belonging to one of the two classes in the dataset. The following equation depicts the possibility of y = 1:

$$P(y = 1|x; w) = \frac{1}{1 + e^{-yw^T x}} \tag{7}$$

where *w* denotes the regression coefficient.

Given this is a minimisation problem, the LR objective function may be stated as:

$$\min_{w,c} \frac{1}{2} w^T w + C \sum_{i=1}^{n} \log\left(\exp\left(-y_i(x_i^T w + c)\right) + 1\right) \tag{8}$$

where *c* is a constant term and *C* is a previously fixed hyperparameter that controls the balance between two terms of the objective function (Hosmer Jr, Lemeshow et al. 2013).

## 6.2. K-Nearest Neighbour

As a simple ML algorithm, the K-Nearest Neighbour (KNN) is a non-parametric supervised classification method. The model is a type of non-generalising learning or instance-based learning that, instead of constructing an internal model, only remembers instances of the training data. A simple majority vote of the nearest neighbours of each sample vector is the basis of this algorithm.

By computing the average observations within a particular neighbourhood, KNN naturally models the connection between the independent factors and the continuous result. Cross-validation may be used to establish or to choose the neighbourhood size to decrease the mean square error. A KNN model is simple to build and may effectively handle non-linearity. Additionally, model fitting is quick since no particular parameter or number has to be computed. KNN may be carried out in three distinct steps:

Step 1: Calculate the distance between the objects to be categorised and calculate their neighbours utilising one of the available distance functions (e.g. Minkowski, Manhattan, Euclidean).

Step 2: Choose the K closest data points (the items with the K lowest distances).

Step 3: Perform a "majority vote" among the data points; the categorisation that receives the most votes in that pool becomes the final categorisation.

Both data inspection and hyper-parameter methods may be used to determine the optimum value for K (Mucherino, Papajorgji et al. 2009, Cheng and Hoang 2016).

### 6.3. Decision Tree

A decision tree model is a tree data structure with an unspecified number of nodes and branches. Internal nodes are those that have outward edges, whereas others are referred to as leaves. A specific internal node divides the examples utilised for classification or regression into two or more categories. During the training phase, the input variable values are compared to a specific function. A decision tree stimulant is an algorithm that generates a decision tree given particular occurrences. The performed algorithm attempts to find the optimal decision trees by lowering the fitness function. Since the datasets of this study contain two classes (stable or failed), a categorisation model is fitted to the target variable employing each independent variable in this study. A decision node is composed of at least two branches. A leaf node denotes a categorisation or decision. The root node is the highest decision node in a tree that correlates to the best predictor.

An overly big tree improves the chance of overfitting the training data and performing poorly on new samples. The sample space may include important structural information, but a too-small tree may not be able to capture it. The horizon effect is a term that refers to this problem. One commonly used strategy is to continue expanding the tree until each node has a limited number of instances and then prune the tree to eliminate nodes with no further information. The primary goal of pruning is to minimise the size of a decision tree while maintaining prediction accuracy as evaluated by a set of cross-validations (Feng, Li et al. 2018).

**6.4. Support Vector Classifier**

Support Vector Machine (SVM) is often used for three primary objectives as a supervised learning technique: classification, regression and pattern recognition. We call an SVM used for classification as a Support Vector Classifier (SVC). Fast computation to achieve globally optimal and effective performance to resist overfitting are two significant characteristics distinguishing SVM from basic artificial neural networks. Considering a dataset with N samples and a represented result type of y (1 or 0), the following training vectors are used for binary classification:

$$D = (x_1, y_1), (x_2, y_2), \ldots, (x_N, y_N) \quad x \in R^n, \quad y_i \in {0, 1} \qquad (9)$$

where n influencing features exist in the n-dimensional space that may be used for addressing the decision boundary. Soft-margin SVMs impose penalties and may be utilised well for nonlinear classification when combined with the kernel technique. Thus, every function f(x) in the SVM kernel trick may be shown in the following manner (Noble 2006):

$$f(x) = w^T \phi(x) + b = 0 \qquad (10)$$

where W is the vector of the output layer, and b denotes the bias. The input variable x is defined as an $N \times n$ matrix, and $\phi(x)$ is a kernel function.

A problem of reduction is proposed to determine w and b (Cortes and Vapnik 1995):

$$\text{Minimise: } \frac{1}{2} w^T + C \sum_{i=1}^{n} \chi_i \qquad (11)$$

$$\text{Subjected to: } y_i(w \cdot x_i + b) \geq 1 - \chi_i$$

Here, C denotes a penalty coefficient while $\chi_i \geq 0$ denotes slack variables, which relate to the misclassification effects.

**6.5. Random Forest**

Random Forest (RF) is a method for ensemble learning based on divide-and-conquer widely used in real-world classification and regression. It uses bootstrap aggregation, commonly known as bagging, to create an ensemble of randomly generated base learners (unpruned DTs). The RF critical point is the creation of a collection of DTs subjected to controlled alternation. Another critical element of the training process is the random choice of features. The method constructs a subset of M important variables for each node in the trees. It is necessary to choose a variety of influencing factors to guarantee DT variation. As a result, each of the DTs conducts an independent assessment of the forest. The results are obtained by voting on the whole tree. In general, the RF algorithm is composed of the following four steps:

1- Randomly choose samples from a dataset.

2- For each sample, construct a decision tree and obtain prediction results from the decision tree.

3- Vote on the outcome of each forecast.

4- Assign the final prediction to the most chosen prediction result.

Additional information regarding RF may be found in References (Belgiu and Drăguţ 2016, Kardani, Bardhan et al. 2021), Kardani, Bardhan et al. 2021) .

**6.6. Naive Bayes**

Naive Bayes (NB) is a basic, but robust, method for classifying binary and multi-class issues. Because NB makes significant assumptions about the independence of characteristics based on Bayes' theory, it may be classified as a probabilistic category. The following equation expresses Bayes' theory:

$$P(A|B) = \frac{P(B|A)P(A)}{P(B)} \qquad (12)$$

where P(A|B) is the possibility of A if B occurs (or posterior probability), viz., a conditional probability implying that A's possibility is dependent on what happens to B.

Gaussian Naive Bayes (Gaussian NB) is an enhanced version of the basic and most common Naive Bayes. The Gaussian distribution requires the standard and mean deviation to be calculated using the training data (Rish 2001) (Tsangaratos and Ilia 2016).

**6.7. Stacking ensemble**

Stacking generalisation is an ensemble algorithm that employs a meta-learning model to learn how to combine the predictions of multiple base models maximising the performance of the combined model. The generalisation method is indeed based on a new machine learning model in the ensemble that learns when to use each base model for a problem. The base models are typically different and are combined by a single meta-model. With different base models having various types of skills on the dataset, the errors in the prediction of different base models are less correlated. When training the meta-model, the base models are fed with out-of-sample data, i.e., the data not already used to train the base models. The predictions made by base models, along with the expected predictions, create the input and output data to train the meta-model.

The level-1 or base models can be diverse and complex where the difference of the approaches and assumptions in base models can help the meta-model see the data in different ways. Here we use a range of various machine learning approaches for the base models including logistic regression (LR), K-Nearest Neighbours (KNN), Decision Tree (DT), Support Vector

Classification (SVC), Random Forest (RF) and Gaussian Naive Bayes (NB). 10-fold cross-validation of the base models is used where the out-of-fold data create the training dataset for the meta-model.

A simple model will suit the level-0 or beta-model which provides a smooth combination of the predictions. Thus, linear models are often used for beta-models. A logistic regression model is selected here for the classification task performed by the beta-model. As such, the beta-model derives a weighted average or blending of the predictions made by the base models. 5-fold cross-validation is used for the beta-model.

While the stacking ensemble is designed to improve the overall performance of model predictions compared to the base models, it is not guaranteed to always achieve an improvement. Different aspects may affect the ensemble performance including the sufficiency of the training data to represent the complexity of the problem and if there are more complex insights not captured by base models to be revealed by the ensemble (Džeroski and Ženko 2004, Rokach 2010, Kardani, Zhou et al. 2021).

## 6.8. Bagging ensemble

Bagging ensemble also known as bootstrap aggregation is one of the ensemble models that combines multiple learners trained on subsamples of the same dataset. Bagging ensemble reduces the variance of the prediction errors and generally improves the accuracy of the model particularly when perturbing the learning dataset can cause a significant change in the predictor constructed.

The bagging ensemble includes a process of (i) creating multiple datasets sampled from the original data, (ii) training multiple learners on each dataset, and (iii) combining all learners to generate a single prediction. The combined prediction is generated by averaging the results of the applied models for a regression analysis or majority voting approach for classification problems.

Sampling with replacement is the core idea in bagging ensembles so that some instances are included in samples multiple times while others are left out. The ensemble can consist of a single base learner invoked several times using different subsets of the training set. A random forest regressor is employed here as the base model (Rokach 2010, (Rokach 2010, Bühlmann 2012).

### 6.9. Voting ensemble (VO En)

As a meta-algorithm, the voting ensemble combines several base models either as regressors or classifiers and produces predictions based on voting algorithms. The ensemble is ideally developed to integrate the predictions of conceptually different base models. A voting ensemble can be beneficial for a set of equally well-performing models to balance out their individual weaknesses.

For regression predictions, the algorithm involves calculating the average of the predictions from base models. In the case of classifier ensembles, a voting ensemble contains several base classifiers. The voting ensemble can use the majority voting (hard voting) in which each base classifier is entitled to one vote where a class with the highest number of votes is determined as the final prediction. If the number of classifiers is even, a tiebreak rule should be identified. The ensemble can also use the average predicted probabilities (soft voting) to combine the predicted classes of base models. The number of base classifiers can be arbitrary (Rokach 2010). In this study, the base models of the VO En are LR, KNN, DT, RF, SVC and Gaussian NB.

### 7. Results and discussion

## 7.1. Metrics

In this study, the accuracy (ACC), F1-score, AUC-score and some metric values extracted from the confusion matrix have been used to compare the performance of the ML models as classifiers for the slope stability problem.

Confusion matrix (see Table 3), also known as the error matrix, is a tabular summary used to describe and visualise the performance of ML models in a supervised classification.

**Table 3** Confusion Matrix

| Actual Values | Predicted Values | |
|---|---|---|
| | Negative | Positive |
| Negative | True Negative, TN | False Positive, FP |
| Positive | False Negative, FN | True Positive, TP |

According to Table 3, some metric values including accuracy, sensitivity (true positive rate), specificity (true negative rate), and false positive rate can be defined. The following equations indicate the mathematic expressions of these metrics:

$$\text{Accuracy (ACC)} = \frac{TP + TN}{\text{total instances}} \quad (13)$$

$$\text{Sensitivity (recall or true positive rate)} = \frac{TP}{TP + FN} \quad (14)$$

$$\text{Specificity (true negative rate)} = \frac{TN}{TN + FP} \quad (15)$$

$$\text{False positive rate} = 1 - \text{Specificity} = \frac{FP}{TN + FP} \quad (16)$$

Another performance metric used in this research is F1-Score, also known as F measure. It is employed as a metric to evaluate the accuracy using both precision ($\frac{TP}{TP+FP}$) and recall ($\frac{TP}{TP+FN}$).

In other words, F1-Score reaches its best value at 1 and worst at 0 through the harmonic average of the precision and recall as expressed in the following equation:

$$F_1 = \frac{TP}{TP + \frac{1}{2}(FP + FN)} \qquad (17)$$

It is worth mentioning that in statistical analysis of geotechnical stability problems, the class of instability events can be rare in comparison with the stable cases. In such problems, the issue of class imbalance should be carefully considered in training the ML models to improve their performance. For uneven class distributions, where a balance between precision and recall is needed, F1-Score can be a better measure to use than the accuracy.

In addition, the receiver operating characteristic (ROC) curve, which plots the true positive rate against the false positive rate at various cut-off values, is used for further analysis of the models. This graphical plot can be employed to express the predictive ability of a model. The area under the ROC curve, i.e., AUC, is commonly used as a measure of the performance of classifiers. AUC ranges in value from 0 to 1, representing a totally wrong to fully correct prediction by a classifier, respectively. As a measure to evaluating the AUC-Score values, one may consider the following guide: Outstanding (0.9 < AUC < 1), Excellent (0.8 < AUC < 0.9) and Acceptable (0.7 < AUC < 0.8) (Hosmer Jr, Lemeshow et al. 2013).

### 7.2. The performance of classifiers on slope stability MC data

*Confusion matrices:* The confusion matrices of RF, SVC and BG En models on test datasets for a single iteration of model training are shown in Table 4. The models are trained on %5 of MC data and tested on the remaining %95. The table provides information on the share of failure and stable classes in each dataset and the model performances on predicting each class. In the $U_{0.1,1}$ dataset, which is best predicted, while the accuracy reaches %96.4 for all RF, SVC and BG En, the incorrectly classified cases are mostly the stable slopes wrongly classified as failed (%3.3 of samples). When the accuracy declines to the lowest value among datasets in the $U_{0.5,25}$ dataset (ACC=0.671 for RF), the wrongly classified classes are both from actually stable and failed slopes with %14.3 and %18.5 of all samples being wrongly classified as failed and stable, respectively. For the entire MC dataset with 119,500 test samples, the models trained only on 500 samples provide an accuracy reaching %84.3. The RF and BG En classifiers wrongly classify only about %8 of all samples in each class of stable and failed slopes.

**Table 4**. Confusion matrices of classifiers on the test datasets. Models are trained on 500 samples where the test datasets consist of 9500 samples for individual U datasets and 119,500 samples for the entire data. The values in parentheses are the numbers normalised over the total number of samples in each test dataset.

| | | RF | | | SVC | | | BG En | | |
|---|---|---|---|---|---|---|---|---|---|---|
| | | Predicted | | Accuracy (ACC) | Predicted | | Accuracy (ACC) | Predicted | | Accuracy (ACC) |
| | Actual | Stable | Failed | | Stable | Failed | | Stable | Failed | |
| $U_{0.1,1}$ | Stable | 7577 (0.798) | 314 (0.033) | 0.964 | 7577 (0.798) | 314 (0.033) | 0.964 | 7578 (0.798) | 313 (0.033) | 0.964 |
| | Failed | 32 (0.003) | 1577 (0.166) | | 27 (0.003) | 1582 (0.167) | | 28 (0.003) | 1581 (0.166) | |
| $U_{0.1,6}$ | Stable | 7640 (0.804) | 409 (0.043) | 0.938 | 7585 (0.798) | 464 (0.049) | 0.949 | 7626 (0.803) | 423 (0.045) | 0.939 |
| | Failed | 184 (0.019) | 1267 (0.133) | | 19 (0.002) | 1432 (0.151) | | 156 (0.016) | 1295 (0.136) | |
| $U_{0.1,12}$ | Stable | 7603 (0.800) | 441 (0.046) | 0.931 | 7555 (0.795) | 489 (0.051) | 0.942 | 7594 (0.799) | 450 (0.047) | 0.933 |
| | Failed | 211 (0.022) | 1245 (0.131) | | 61 (0.006) | 1395 (0.147) | | 184 (0.019) | 1272 (0.134) | |
| $U_{0.1,25}$ | Stable | 7612 (0.801) | 413 (0.043) | 0.931 | 7544 (0.794) | 481 (0.051) | 0.942 | 7610 (0.801) | 415 (0.044) | 0.935 |
| | Failed | 242 (0.025) | 1233 (0.130) | | 71 (0.007) | 1404 (0.148) | | 205 (0.022) | 1270 (0.134) | |
| $U_{0.3,1}$ | Stable | 6530 (0.687) | 342 (0.036) | 0.886 | 6580 (0.693) | 292 (0.031) | 0.888 | 6578 (0.692) | 294 (0.031) | 0.889 |
| | Failed | 744 (0.078) | 1884 (0.198) | | 773 (0.081) | 1855 (0.195) | | 762 (0.080) | 1866 (0.196) | |
| $U_{0.3,6}$ | Stable | 6208 (0.653) | 772 (0.081) | 0.823 | 5986 (0.630) | 994 (0.105) | 0.818 | 6196 (0.652) | 784 (0.083) | 0.827 |
| | Failed | 908 (0.096) | 1612 (0.170) | | 738 (0.078) | 1782 (0.188) | | 859 (0.090) | 1661 (0.175) | |
| $U_{0.3,12}$ | Stable | 6417 (0.675) | 775 (0.082) | 0.795 | 6368 (0.670) | 824 (0.087) | 0.800 | 6458 (0.680) | 734 (0.077) | 0.803 |
| | Failed | 1168 (0.123) | 1140 (0.120) | | 1075 (0.113) | 1233 (0.130) | | 1137 (0.120) | 1171 (0.123) | |
| $U_{0.3,25}$ | Stable | 6387 (0.672) | 742 (0.078) | 0.785 | 6536 (0.688) | 593 (0.062) | 0.789 | 6337 (0.667) | 792 (0.083) | 0.788 |
| | Failed | 1296 (0.136) | 1075 (0.113) | | 1407 (0.148) | 964 (0.101) | | 1226 (0.129) | 1145 (0.121) | |
| $U_{0.5,1}$ | Stable | 4729 (0.498) | 582 (0.061) | 0.853 | 4640 (0.488) | 671 (0.071) | 0.853 | 4737 (0.499) | 574 (0.060) | 0.854 |
| | Failed | 812 (0.085) | 3377 (0.355) | | 721 (0.076) | 3468 (0.365) | | 812 (0.085) | 3377 (0.355) | |
| $U_{0.5,6}$ | Stable | 3766 (0.396) | 1223 (0.129) | 0.721 | 3725 (0.392) | 1264 (0.133) | 0.719 | 3759 (0.396) | 1230 (0.129) | 0.723 |
| | Failed | 1432 (0.151) | 3079 (0.324) | | 1403 (0.148) | 3108 (0.327) | | 1405 (0.148) | 3106 (0.327) | |
| $U_{0.5,12}$ | Stable | 3710 (0.391) | 1399 (0.147) | 0.682 | 3611 (0.380) | 1498 (0.158) | 0.679 | 3607 (0.380) | 1502 (0.158) | 0.683 |
| | Failed | 1619 (0.170) | 2772 (0.292) | | 1553 (0.163) | 2838 (0.299) | | 1508 (0.159) | 2883 (0.303) | |
| $U_{0.5,25}$ | Stable | 3883 (0.409) | 1426 (0.150) | 0.667 | 3862 (0.407) | 1447 (0.152) | 0.668 | 3948 (0.416) | 1361 (0.143) | 0.671 |
| | Failed | 1734 (0.183) | 2457 (0.259) | | 1708 (0.180) | 2483 (0.261) | | 1760 (0.185) | 2431 (0.256) | |
| **Entire data** | Stable | 75255 (0.630) | 9548 (0.080) | 0.833 | 77725 (0.650) | 7078 (0.059) | 0.821 | 75379 (0.631) | 9424 (0.079) | 0.834 |
| | Failed | 10408 (0.087) | 24289 (0.203) | | 14318 (0.120) | 20379 (0.171) | | 10416 (0.087) | 24281 (0.203) | |

***Repeated k-fold cross-validation:*** To assess the sensitivity of the ML model performances, a repeated k-fold cross-validation (CV) scheme is employed in this study. In k-fold CV, the dataset is first randomly divided into k separate folds with the same number of instances. The model is tested over each fold in turn where the other $k-1$ folds are used for training the model. To reduce the variance of the performance estimates, it is recommended to perform repeated k-fold CV (Wong and Yeh 2019). k-fold CV with a large number of folds and a small number of replications is shown to be an appropriate method for the performance evaluation of classification algorithms (Refaeilzadeh, Tang et al. 2009). Three repeats of 10-fold CV create the distribution of performance scores such as accuracy and F1 in our study. For each model, distribution of performance scores with a higher mean value and smaller variances is desirable. The CV scheme adopted in this study is illustrated in Fig. 4.

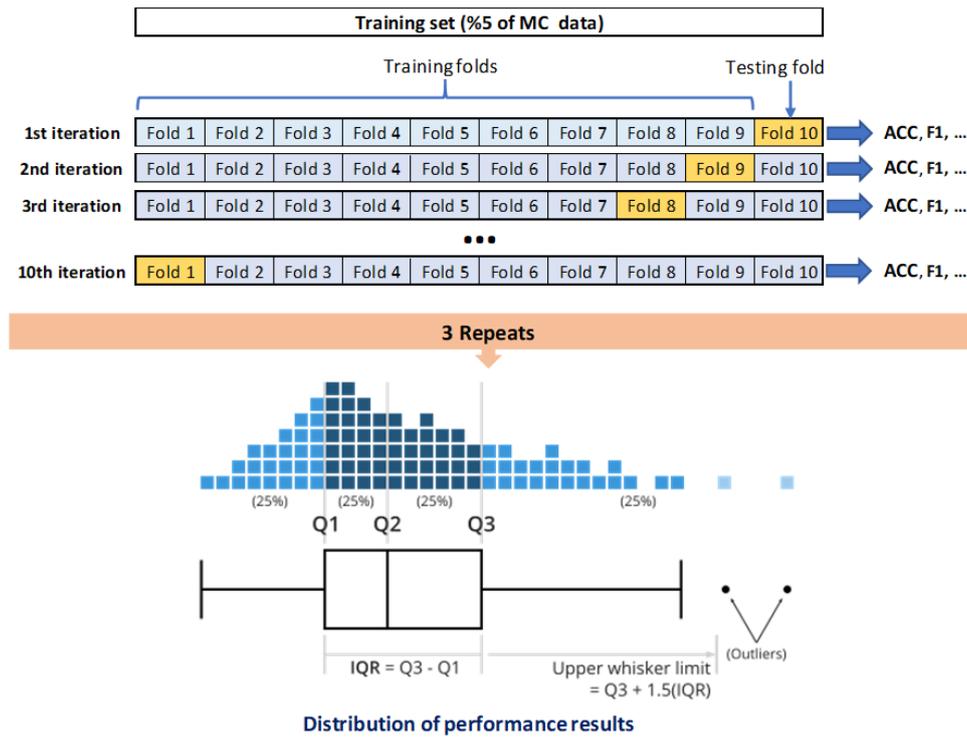

**Fig. 4** 10-fold cross-validation with three repeats adapted in this study resulting in the statistics of the model performance scores shown as boxplots. The interquartile range, IQR is identified by Q1 or $25^{th}$ to Q3 or $75^{th}$ percentile as the boxes. The minimum and maximum are shown as whiskers, and outliers are identified as circles.

***Performance scores and their distributions:*** The mean values of performance results of different classifiers obtained from repeated k-fold CV are summarised in Table . Classifiers are used as MC surrogate models to predict the stability status of heterogenous and anisotropic slopes. For training the surrogate classifiers, %5 of MC data, i.e., 500 samples, are randomly chosen from each random field dataset U containing 10000 samples. The mean performance scores of CV suggest that in general, RF, SVC, and BG En are the most appropriate classifiers for this study. In the dataset where the highest performance scores are achieved ($U_{0.1,1}$ characterised by low heterogeneity and no anisotropy), the BG En model provides the highest scores (F1=0.903, ACC=0.962, AUC=0.98) which are also partly achieved by other models. However, on the most challenging dataset ($U_{0.5,25}$ characterised by high heterogeneity and anisotropy), the RF model reaches the highest ACC=0.691 and AUC=0.721 where the highest F1 score belongs to NB (F1=0.659). On the entire dataset where all sub-datasets are combined, BG En and RF model outperform other models with ACC=0.847, AUC=0.912 and F1=0.737 (Table 5).

In addition to the mean performance scores, the distribution of these scores obtained from three repeats of CV provides more insights into the overall appropriateness of different classification models for the current study. These distributions can better demonstrate the comparison of different models where smaller variance on higher scores will be desirable. The distribution of F1 and accuracy scores for each classification model is shown in **Error! Reference source not found.** as boxplots. In the best-predicted dataset (Fig. 5a: $U_{0.1,1}$), SVC, RF, Bayes, BG En and VO En achieve comparable performance with a similar distribution of F1 and accuracy scores. The similar achievement of different models implies higher predictability of the dataset due to the smaller COV of strength parameter and stronger correlations between mean strength values and the stability status; however, with the same COV, but larger scales of fluctuations ($\xi = 25$), more complex failure mechanisms may have come into play reducing the predictability of the stability. In this dataset (Fig. 5b: $U_{0.1,25}$), SVC model outperforms all models with distributions concentrated at higher values on both ACC and F1 scores. For $U_{0.3,1}$ dataset where COV of shear strength values is increased to %30 (Fig. 5c), SVC results in best scores and small variances, while for the $U_{0.3,25}$ dataset where anisotropy is also significant (Fig. 5d), VO En and RF model appear to be more efficient when both scores are considered. In datasets with the highest COV but isotropic heterogeneity (Fig. 5e: $U_{0.5,1}$), NB and BG En can be selected as the best classifiers. For the same COV but highly anisotropic dataset (Fig. 5f: $U_{0.5,25}$), BG En, RF and SVC perform well in accuracy distributions and NB can be selected as the winner model considering the F1 score.

**Table 5.** Performance metrics of different models and the ensemble classifiers on different random field data. The mean values of the metrics obtained from a repeated 10-fold cross-validation scheme are shown. The highest mean metric values achieved for each random field are shown as bold numbers. 500 samples are randomly taken for model training from each random field dataset.

| Random field | LR | | | KNN | | | DT | | | SVC | | | RF | | | Naive Bayes (Gaussian NB) | | |
|---|---|---|---|---|---|---|---|---|---|---|---|---|---|---|---|---|---|---|
| | F1 | ACC | AUC | F1 | ACC | AUC | F1 | ACC | AUC | F1 | ACC | AUC | F1 | ACC | AUC | F1 | ACC | AUC |
| $U_{0.1,1}$ | 0.878 | 0.955 | 0.976 | 0.862 | 0.950 | 0.976 | 0.786 | 0.931 | 0.814 | **0.903** | **0.962** | 0.977 | 0.901 | **0.962** | 0.973 | **0.903** | **0.962** | 0.976 |
| $U_{0.1,25}$ | 0.666 | 0.899 | **0.967** | 0.800 | 0.931 | 0.959 | 0.645 | 0.884 | 0.806 | **0.820** | **0.936** | 0.955 | 0.775 | 0.924 | **0.961** | 0.662 | 0.848 | 0.947 |
| $U_{0.3,1}$ | 0.718 | 0.843 | 0.915 | 0.712 | 0.834 | 0.906 | 0.632 | 0.799 | 0.765 | 0.751 | **0.880** | 0.922 | 0.746 | 0.877 | **0.940** | 0.712 | 0.807 | 0.893 |
| $U_{0.3,25}$ | 0.429 | 0.729 | 0.742 | 0.451 | 0.759 | 0.755 | 0.430 | 0.693 | 0.643 | 0.455 | 0.775 | 0.782 | 0.508 | 0.775 | 0.812 | **0.568** | 0.699 | 0.772 |
| $U_{0.5,1}$ | 0.762 | 0.793 | 0.863 | 0.738 | 0.711 | 0.846 | 0.634 | 0.681 | 0.678 | 0.816 | 0.842 | 0.922 | 0.808 | **0.843** | 0.922 | **0.823** | 0.842 | 0.909 |
| $U_{0.5,25}$ | 0.537 | 0.583 | 0.658 | 0.597 | 0.613 | 0.646 | 0.521 | 0.581 | 0.565 | 0.610 | 0.677 | **0.721** | 0.614 | **0.691** | 0.721 | **0.659** | 0.657 | 0.673 |
| Entire data | 0.603 | 0.772 | 0.781 | 0.619 | 0.824 | 0.787 | 0.583 | 0.758 | 0.697 | 0.701 | 0.845 | 0.885 | 0.722 | **0.847** | 0.908 | 0.643 | 0.751 | 0.794 |

| Random field | Stacking Ensemble | | | Bagging Ensemble | | | Voting Ensemble | | |
|---|---|---|---|---|---|---|---|---|---|
| | F1 | ACC | AUC | F1 | ACC | AUC | F1 | ACC | AUC |
| $U_{0.1,1}$ | 0.901 | 0.961 | 0.975 | **0.903** | **0.962** | **0.980** | 0.903 | 0.961 | 0.975 |
| $U_{0.1,25}$ | 0.789 | 0.927 | 0.961 | 0.778 | 0.926 | 0.959 | 0.791 | 0.929 | 0.964 |
| $U_{0.3,1}$ | 0.751 | 0.877 | 0.937 | **0.754** | 0.873 | **0.940** | 0.744 | 0.871 | 0.933 |
| $U_{0.3,25}$ | 0.420 | 0.773 | 0.781 | 0.499 | **0.780** | 0.813 | 0.478 | **0.780** | 0.811 |
| $U_{0.5,1}$ | 0.816 | 0.841 | 0.903 | 0.816 | **0.843** | **0.923** | 0.820 | 0.841 | 0.914 |
| $U_{0.5,25}$ | 0.611 | 0.669 | 0.681 | 0.617 | 0.681 | 0.711 | 0.606 | 0.673 | 0.708 |
| Entire data | 0.717 | 0.844 | 0.878 | **0.737** | 0.841 | **0.912** | 0.696 | 0.844 | 0.884 |

The efficiency of ML classifiers when implemented on the entire MC dataset combining various datasets with different COVs and ξ values is also examined (Fig. 7a and b). For this case, a training dataset of 500 samples (0.42% of data) is used. Here, RF and BG En appear to be the most efficient models.

In general, based on the distribution of accuracy and F1 scores, one can conclude that SVC, RF and BG En are the most reliable choices for predicting the stability status of heterogeneous and anisotropic slopes.

Based on AUC scores, for the dataset of low heterogeneity and no anisotropy (Fig. 6a: $U_{0.1,1}$), BG En achieves the highest score (AUC=0.980). For low heterogeneity but high anisotropy (Fig. 6b: $U_{0.1,25}$), LR provides the best score as AUC=0.967. However, for both datasets, all classifiers perform relatively well denoting the high predictability of the datasets except the DT classifier which can be deemed inappropriate for this problem. With higher heterogeneity, regardless of the anisotropy levels, RF and BG En provide the highest scores (AUC=0.94 for $U_{0.3,1}$ and AUC=0.813 for $U_{0.3,25}$, Fig. 6c and d). BG En also performs best for the highest heterogeneity but no anisotropy (AUC=0.923, Fig. 6e) whereas RF outperforms others with AUC=0.721 for the highest heterogeneity and anisotropy (Fig. 6f). When the entire MC dataset is examined, RF and BG En are the most appropriate classifiers with AUC=0.89 (Fig. 7c).

In general, based on the AUC metric, RF and BG En can be concluded as the most appropriate classifiers for this problem where DT seems inappropriate.

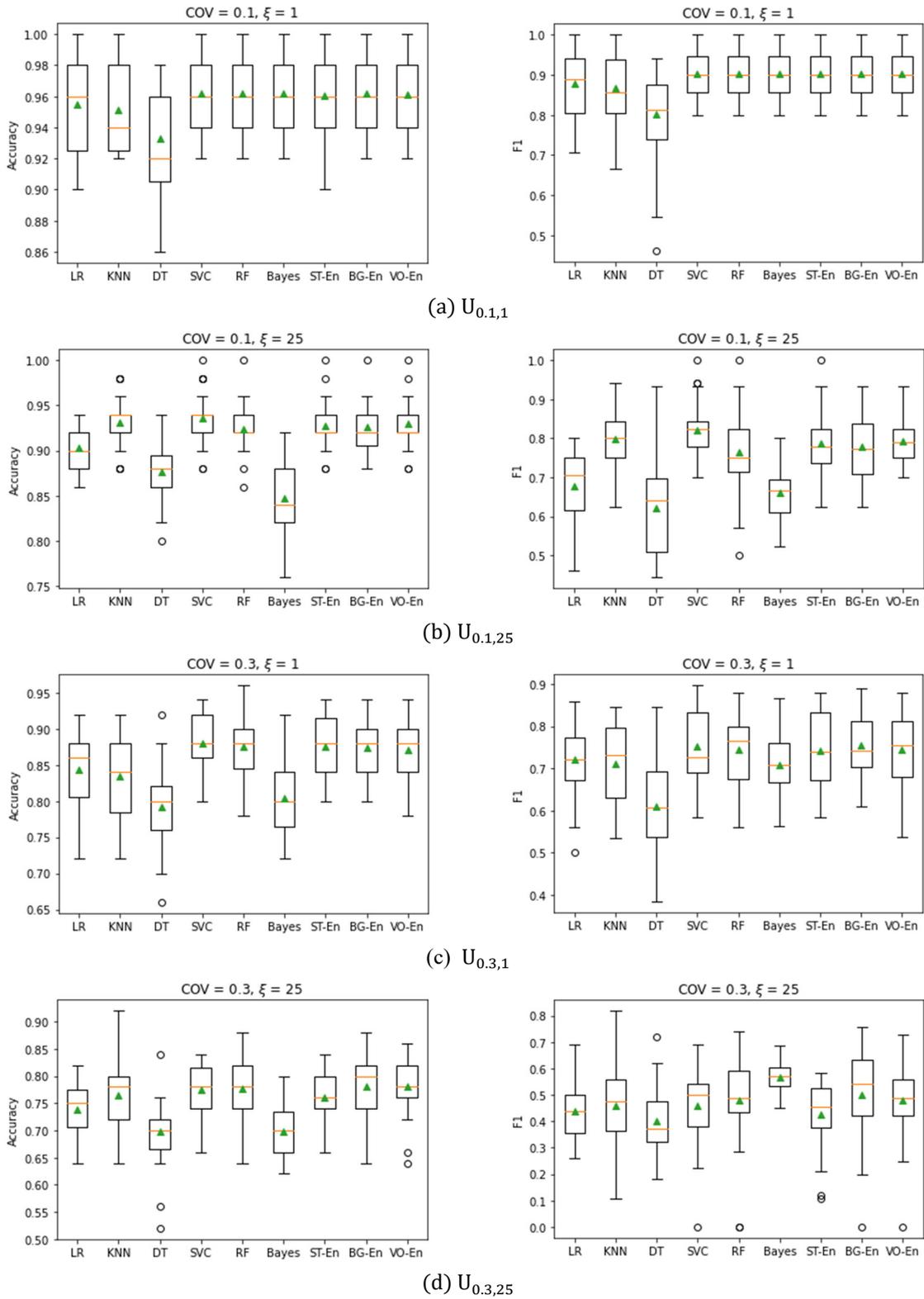

**Fig. 5.** The performance of different algorithms obtained from three repeats of 10-fold cross-validation. Boxplots show the mean (green triangle), median (yellow line), interquartile range (25th to the 75th percentile as the boxes), minimum and maximum (whiskers), and outliers (circles) for accuracy scores (left panels) and F1 (right panels) for each model. Models are trained on 500 samples from each random field dataset.

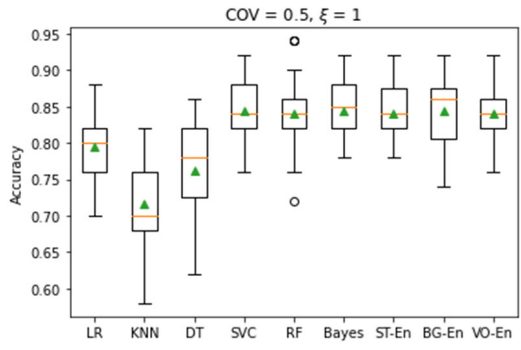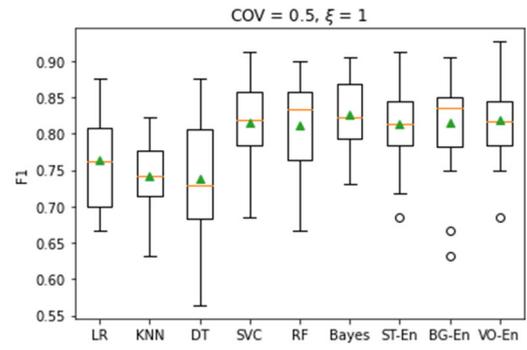

(e) $U_{0.5,1}$

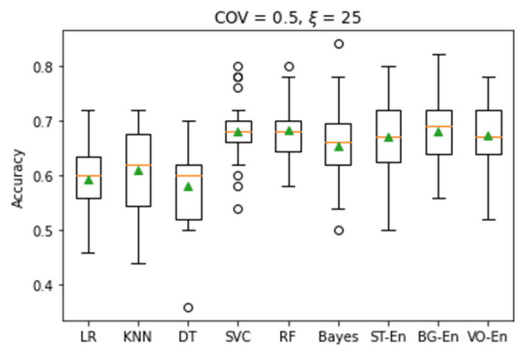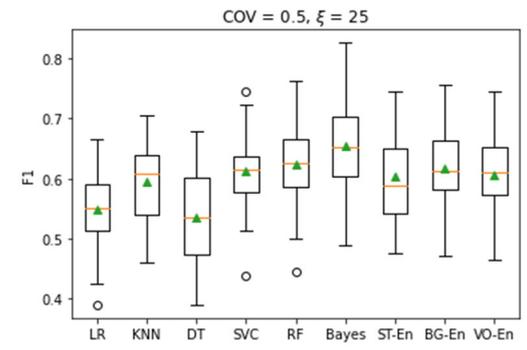

(f) $U_{0.5,25}$

**Fig. 5** Continued.

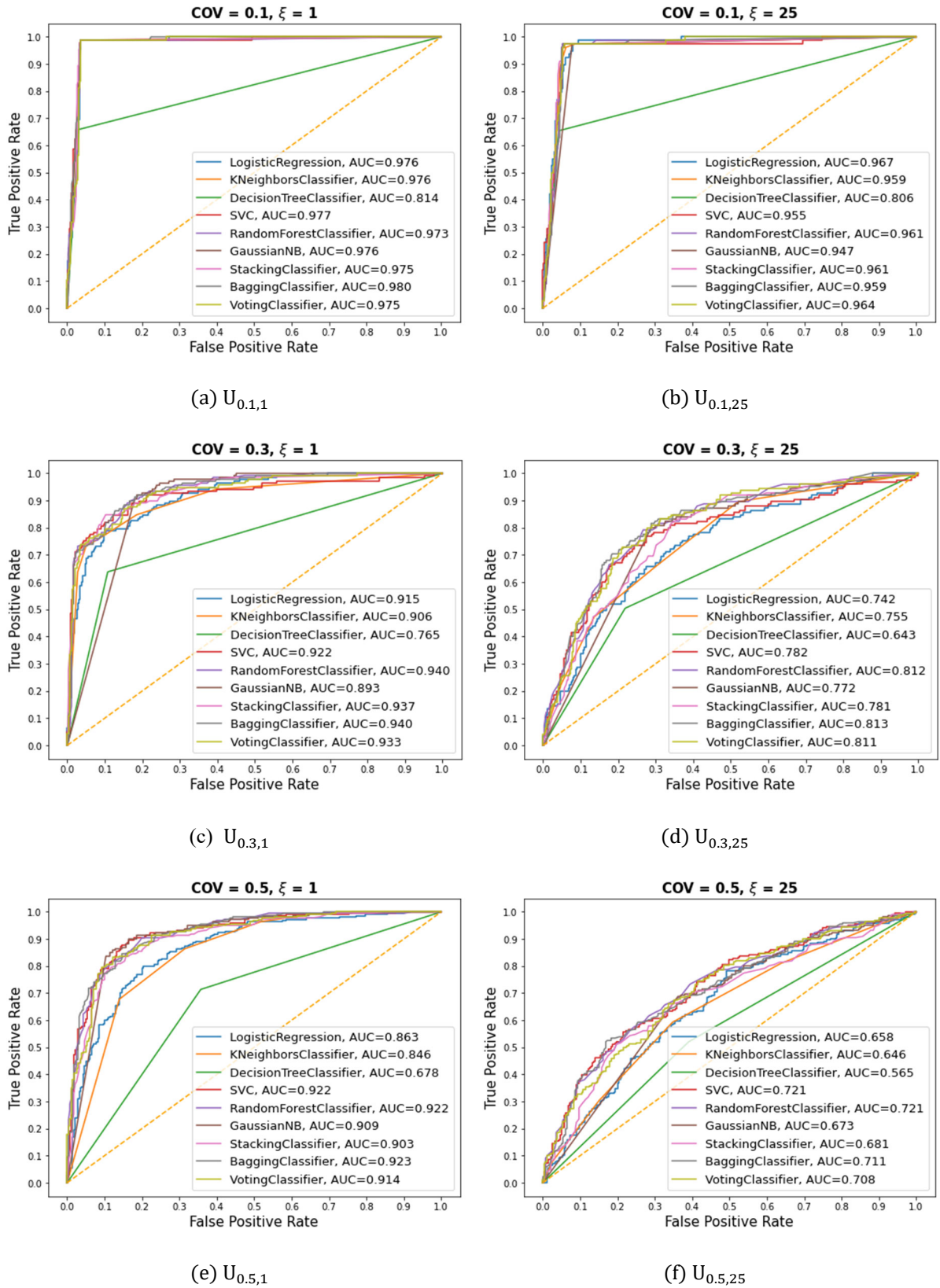

**Fig. 6** AUC ROC curves for different machine learning algorithms on various datasets. Models are trained on 500 samples from each random field dataset.

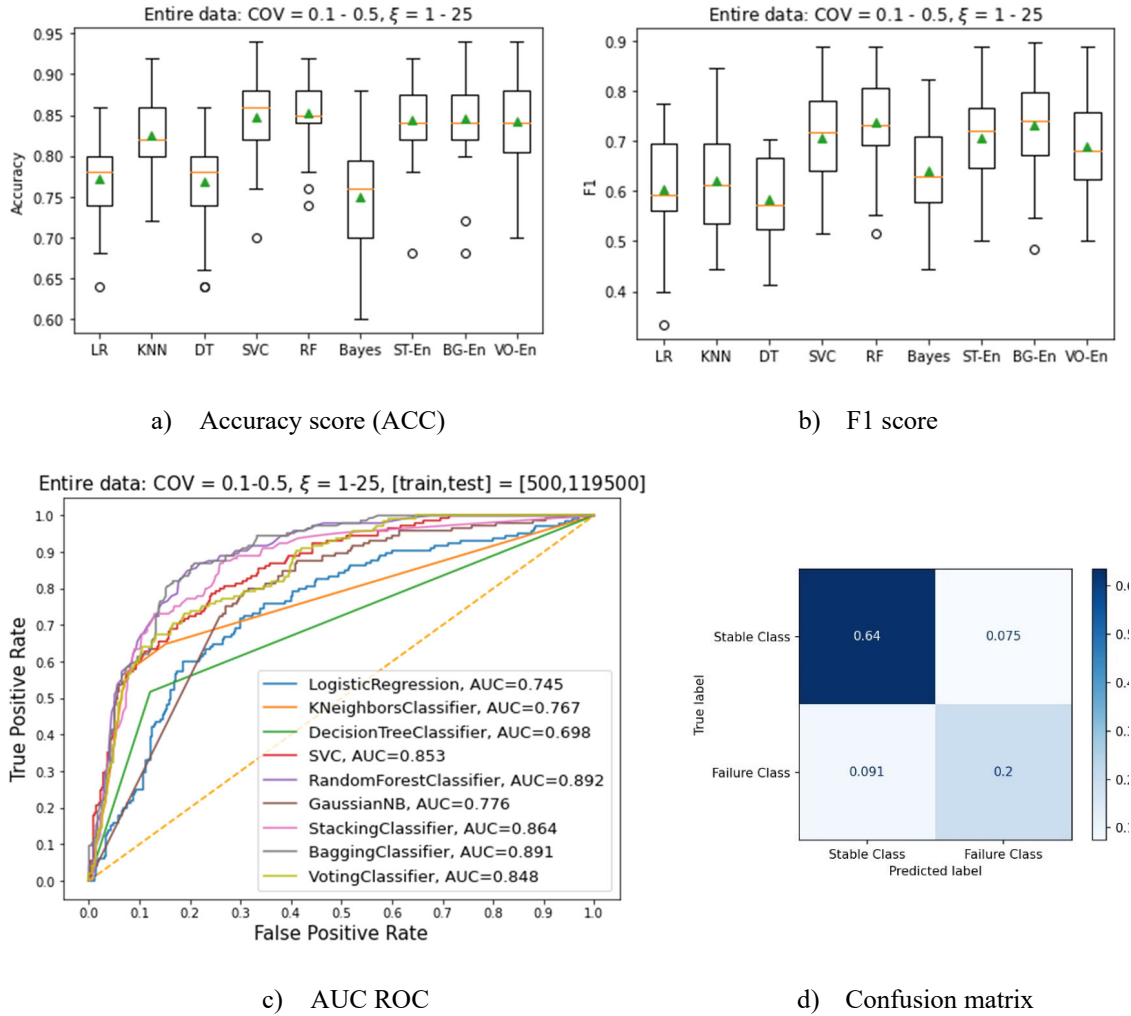

a) Accuracy score (ACC)   b) F1 score

c) AUC ROC   d) Confusion matrix

**Fig. 7** The performance of different algorithms on the entire random field data (all COV and ξ datasets). Results are obtained from three repeats of 10-fold cross-validation. The models are trained on 500 samples (%0.42 of data). The confusion matrix showing the normalised values over the size of the test dataset is shown in d) as the results of the RF model trained with 500 samples for a test against 119,500 samples.

Fig. 8 provides an overview of the performance of different classifiers on different datasets characterised by various levels of soil heterogeneity and anisotropy. With F1 and AUC metrics being the most important measures to evaluate binary classifiers in unbalanced classes, the graphs of Fig. 8 present the overall performance of models with combined metrics. In the case of random field data with no anisotropy ($\xi = 1$), regardless of the level of soil heterogeneity, the BG En appears to outperform other classifiers with higher F1 and AUC scores (Fig. 8a, c and e). The RF and VO En are also the next appropriate models with performance scores close to the best values. In the case of random field data with high anisotropy ($\xi = 25$), the RF is perhaps the most reliable model while the VO En has performed better for low heterogeneity (Fig. 8b) and SVC has performed similarly well at high heterogeneity (Fig. 8f). Ultimately, BG En outperforms other models when learning the entire MC dataset of different heterogeneity and anisotropy levels (Fig. 9).

In general, based on the combined assessment of F1 and AUC measures for all datasets, we can conclude that BG En and RF are the most appropriate models for this study while SVC can be the runner up model. DT, KNN and LR are the least efficient models for this study where Naive's performance has been inconsistent.

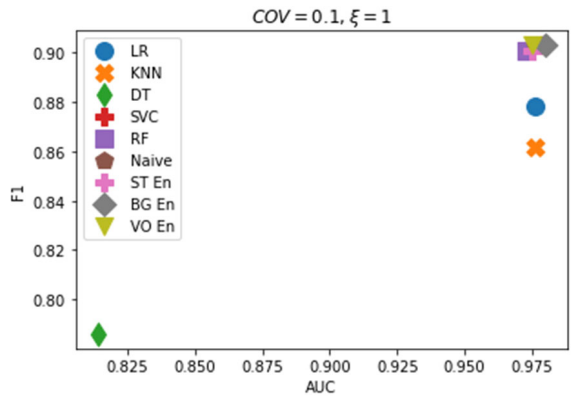

(a) $U_{0.1,1}$

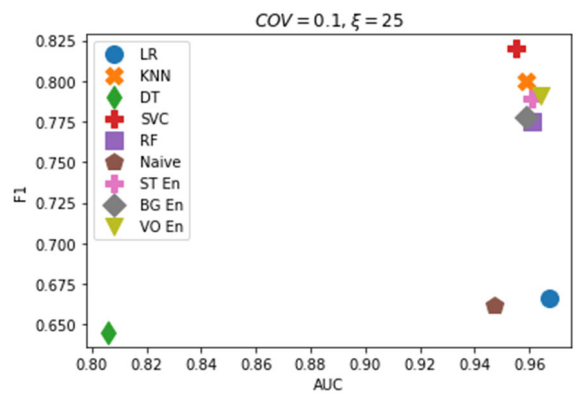

(b) $U_{0.1,25}$

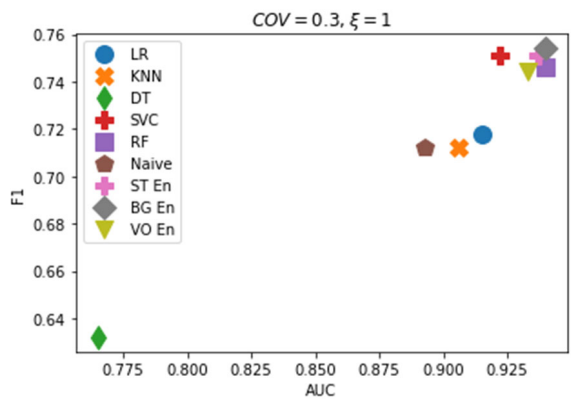

(c) $U_{0.3,1}$

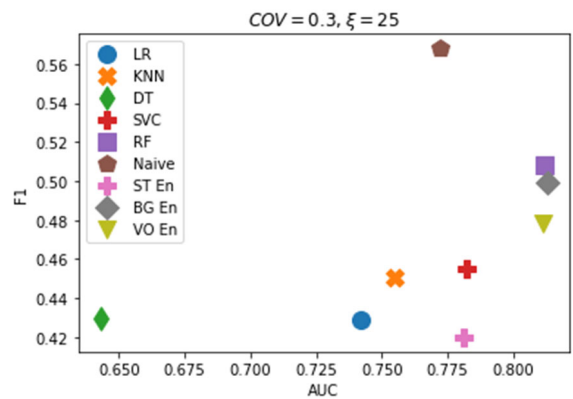

(d) $U_{0.3,25}$

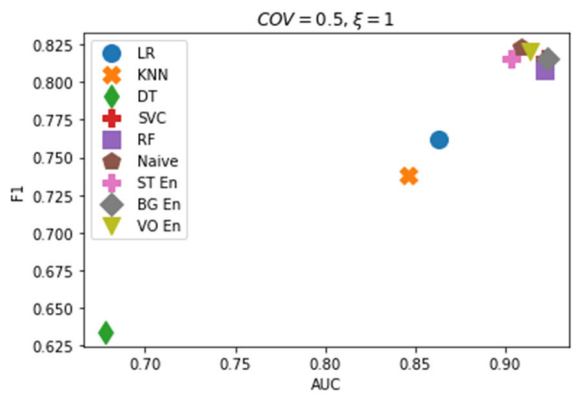

(e) $U_{0.5,1}$

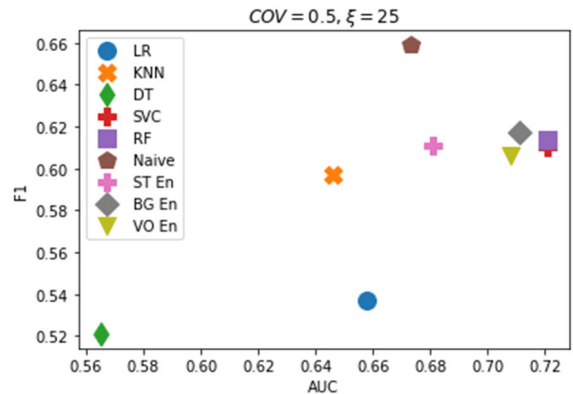

(f) $U_{0.5,25}$

**Fig. 8** AUC versus F1 scores for different algorithms on various datasets trained with 500 samples (%5 of each dataset). Datapoint values are the mean of repeated CV results.

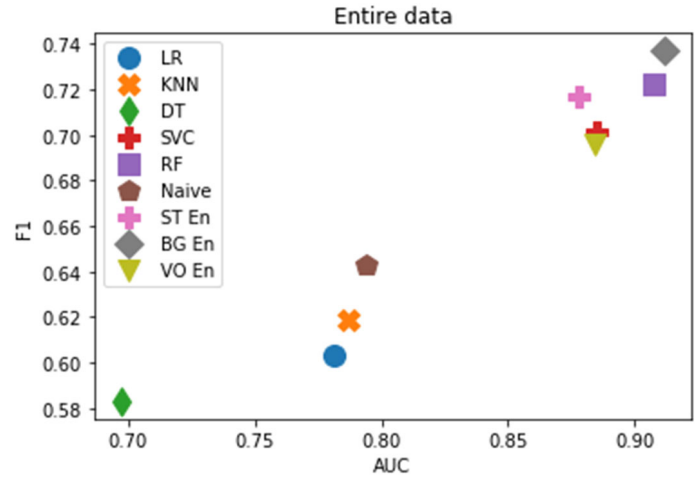

**Fig. 9** AUC versus F1 scores for different algorithms on the entire random field dataset (120,000 samples) trained with 500 samples (%0.42 of data).

## 7.3. The effects of the heterogeneity and anisotropy of random fields on classifier performances

Fig. 10 compares the performance of the selected ML classifiers as surrogate models in reliability analysis with respect to different levels of heterogeneity (COV = $0.1 - 0.5$) and anisotropy ($\xi = 1 - 25$). The best performing classifiers, RF, SVC and BG En are trained on %5 of the data (500 samples) for each dataset where the datapoints in Fig. 10 are the mean values of the results obtained from the three times repeated CV. A general trend in all models is the declining performance (decreased accuracy and AUC scores) with increasing heterogeneity and anisotropy. This behaviour can lie in the increased complexity of the failure mechanisms with these properties of random soils. The failure in a less heterogeneous slope is better correlated with the average of the random strength variables. Such slopes are also more likely to fail via typical shear surfaces which are better learned by the classifiers. A semi-circular geometry of the failure surface is attributed to these failure cases. With increasing heterogeneity, the predictability decreases as the failure mechanisms are more complex depending on the interconnection of local week zones with irregular surfaces of failure.

The effects of anisotropy are also evident in Fig. 10, where a higher anisotropy ratio ($\xi$) leads to lower predictability of the stability status of slopes. These effects could also root in the complexity of the failure mechanisms where not only the local week zones (due to heterogeneity) may cause the failure, but also the correlated week layers distributed spatially into the field can contribute to irregular failure surfaces. Therefore, any failure geometry from a circular deep surface to a local shearing zone or a shallow or deep layering slide can be a possible mechanism. The lower repeatability of different mechanisms can reduce the prediction ability of the models.

In general, RF and BG En perform similarly well on all datasets while according to the AUC metric (Fig. 10b), SVC is less efficient than the other two models in the moderate level of heterogeneity (COV=0.3).

It should be noted that while the F1 score is used as a useful measure to evaluate the performance of classifiers in unbalanced datasets, this score by definition is not comparable on different datasets with varying ratios of unbalanced classes. Thus, the variation of the F1 score is not shown for the comparison of different classifiers in Fig. 10.

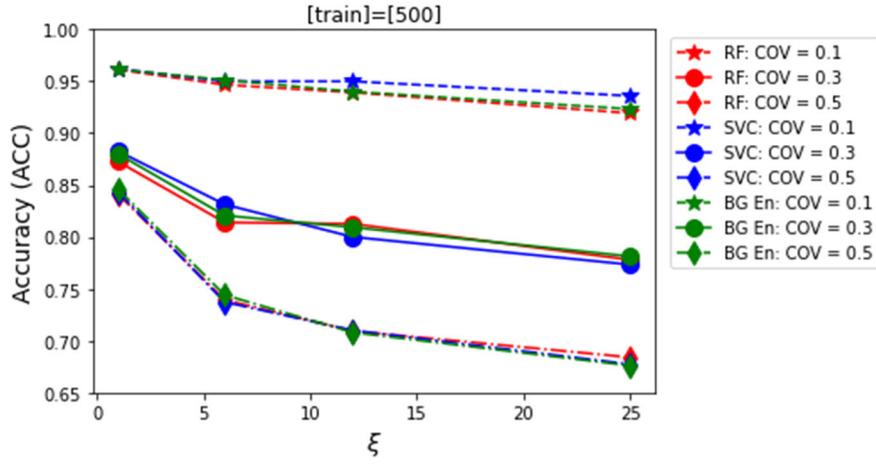

(a)

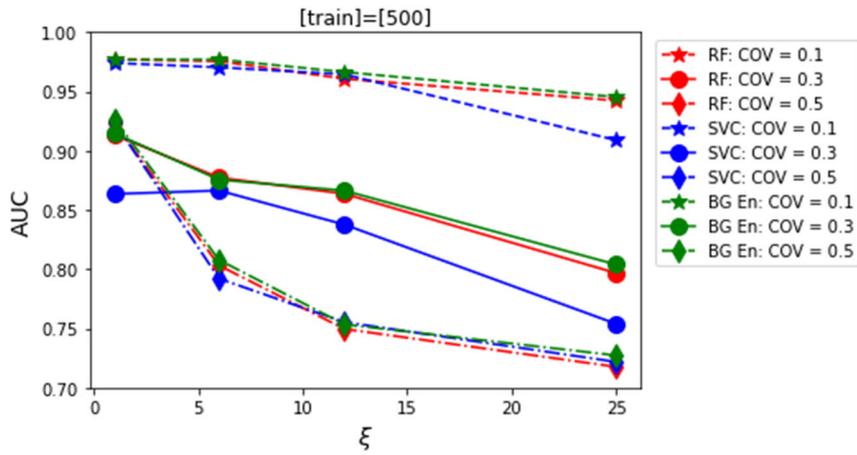

(b)

**Fig. 10** Accuracy Score, ACC (a) and Area Under Curve, AUC score (b) vsversus the anisotropy ratio ξ for Random Forest (RF), Support Vector (SVC) and Bagging ensemble (BG En) trained with 500 samples for each dataset obtained as mean values from 3-times repeated 10-fold CV.

### 7.4. Predicted probability of failure

While the ACC and AUC for the binary classification (failure, non-failure) may decrease to about %70, it should be noted that the errors in the predicted probability of failure, $p_f$ can be in much lower extent. The nature of binary results either as failure or non-failure can induce a high number of near-to-failure samples being incorrectly classified. While in the case of the predictions of FOS data, fluctuations in the near-to-failure (FOS≈1) predictions can induce a smoothed error in the mean FOS calculated, similar model performances on a binary classification may induce lower accuracies. However, when calculating the $p_f$, the ratio as the number of failure predictions over the total number of samples is calculated. Therefore, near-to-failure mis-classified samples can be balanced which can reduce the average errors in the calculated $p_f$.

The probability of failure $p_f$ predicted by ML models trained on only %5 of random field data is shown in Fig. 11. The errors in the predicted $p_f$ range from less than %1 for all models and with an average of 0.46% for all cases of heterogeneity and anisotropy (Fig. 11 -a).

The actual $p_f$ values obtained from the complete MC data are shown as data points in Fig. 11-b. The variation of ML predictions of $p_f$ is shown as the bond of the standard deviations. Trained only on %5 of data, ML surrogate models can provide highly accurate predictions.

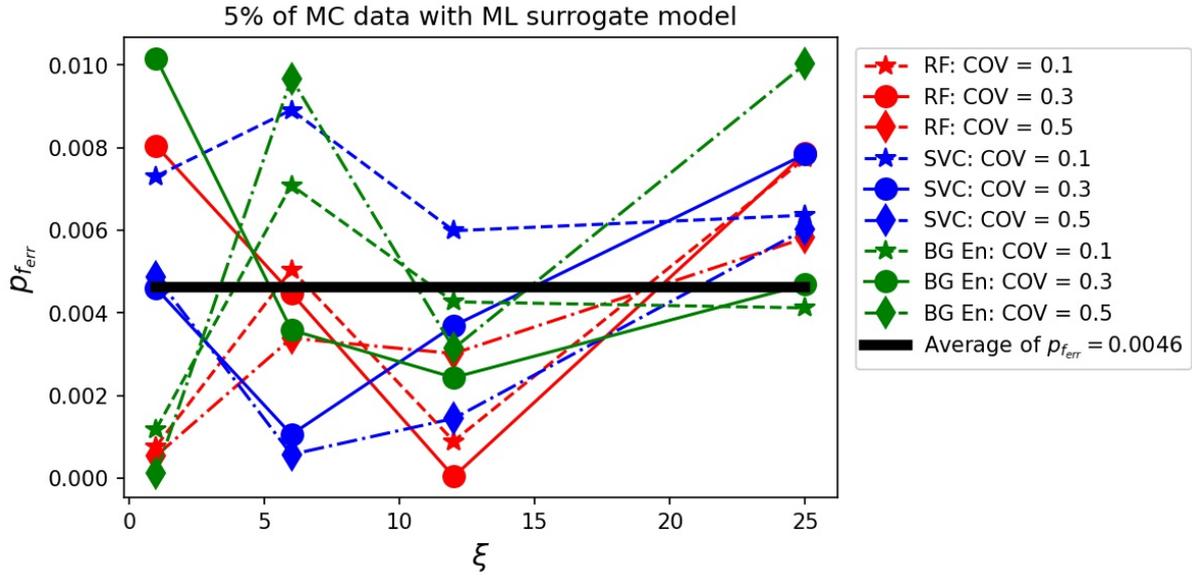

(a)

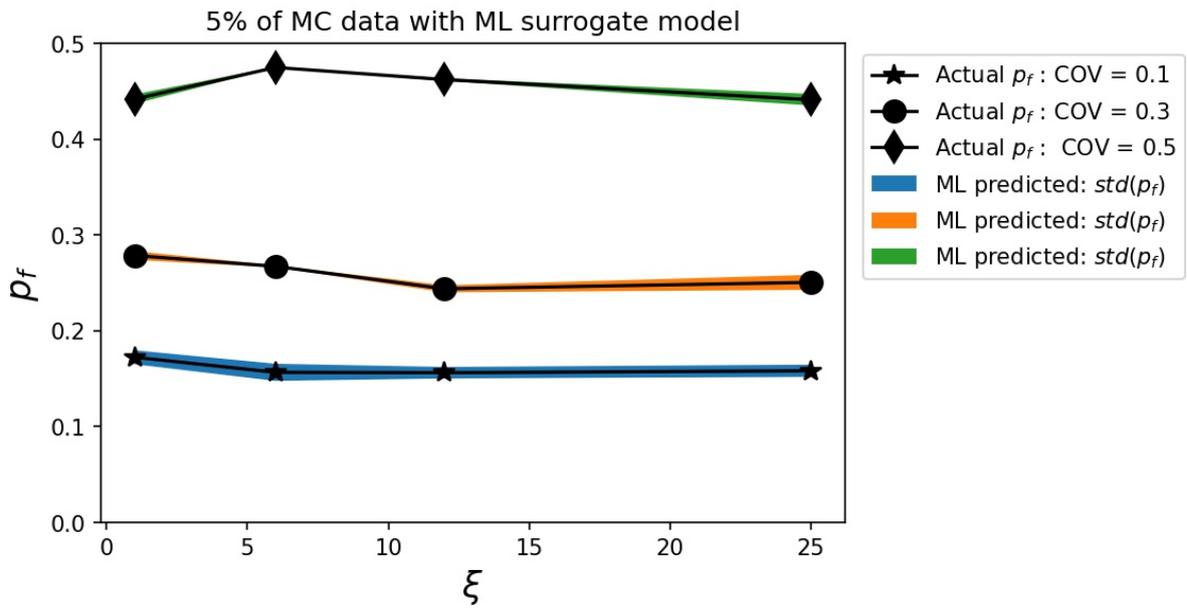

(b)

**Fig. 11** The errors of the ML predicted probability of failure $p_{f_{err}}$ (a) and the standard deviation (std) of ML predicted $p_f$ (b) versus anisotropy ratio $\xi$. The SVC model is used here. The actual $p_f$ in (b) is obtained from complete MC data. ML models are trained with 500 samples for each dataset (%5 of data).

## 7.5. Computational time

The CPU time for each random finite difference simulation without a FOS calculation to determine the stability status of the slopes is 43 seconds on average on a Laptop with Intel Core i3-5010U 2.1GHz processor using four cores. Thus, each dataset of particular ($COV, \xi, \mu_{C_u}$) values consisting of 2,000 simulations takes a CPU time of 23.9 hours to finish. The entire random field MC dataset containing 120,000 simulations is completed in about 60 days. A similar dataset with calculated FOSs can take up to 306 days to complete. The ML models are trained on 500 samples to predict the entire dataset. The computation time for generating this training dataset will take only about 6 hours where the ML training and predictions are performed within a few minutes. Therefore, the proposed ML models can reduce the computational CPU time of such study from 306 days to 6 hours where an accuracy score of ~%85 and AUC score of ~%91 can be expected. In the case of the random finite difference simulations with FOS calculation using strength reduction technique, the CPU time for each simulation can reach 220 seconds on average demanding unaffordable computational costs for such a study. A summary of the computation time and expected performance metrics is presented in Table 6.

**Table 6.** Comparison of the computational time of original and ML aided MC methods

| MC dataset | Original MC method | | | ML aided MC method * | | | | | |
|---|---|---|---|---|---|---|---|---|---|
| | | with FOS calculations | without FOS calculations | | | | | | |
| | MC samples | CPU time (hours) | CPU time (hours) | MC samples | CPU time (hours) | Accuracy (ACC) | F1 | AUC | $P_{f_{err}}$ |
| U datasets (specific COV and ξ) | 10,000 | 611 (~25.5 days) | 119.4 (~5 days) | 500 (%5 of MC data) | 5.97 (~0.25 days) | Min: 0.691– Max: 0.962 | Min: 0.568– Max: 0.903 | Min: 0.721– Max: 0.980 | ≤0.010 |
| Entire dataset | 120,000 | 7333 (~306 days) | 1433 (~60 days) | 500 (%0.42 of MC data) | 5.97 (~0.25 days) | 0.847 | 0.737 | 0.912 | 0.0046 |

* Min and max metric values are associated with datasets $U_{0.5,25}$ and $U_{0.1,1}$, respectively.

## 8. Conclusion and future research direction

In a systematic approach in this study, we investigated the efficiency of ML classifier methods as response/surrogate models to predict the results of the random field slope stability with spatial variability. A complete Monte Carlo database of heterogenous anisotropic slopes containing 120,000 simulations is generated. The random field variable is the soil undrained shear strength with COV ranging from 0.1 to 0.5, and the anisotropy ratio, ξ ranging from 1 to 25. Predictions on the whole database are made by machine learning surrogate models trained on only 500 samples. The findings of this research can be summarised as follows:

- Repeated k-fold cross-validation revealed the sensitivity of ML surrogate models providing the distribution of performance scores. A detailed study suggested that the Bagging ensemble and Random Forest are the most appropriate models for this random field study with SVC being the next appropriate choice. DT, KNN and LR are the least efficient models for this study where Naive's performance is inconsistent.

- The efficiency of ML surrogate models generally decreases with higher heterogeneity and anisotropy of random fields.

- Slightly heterogeneous random fields are highly predictable: when trained on 500 samples (%5 of MC dataset), the accuracy of ML surrogate models for classifying the failure and non-failure cases reached %96.2 for (COV=0.1, $\xi$=1), although decreases to %93.6 for (COV=0.1, $\xi$=25).

- With higher heterogeneity, the classification predictability decreases: the accuracy of ML surrogate models trained on %5 of data is %88 for (COV=0.3, $\xi$=1), and %84 for (COV=0.5, $\xi$=1).

- A higher anisotropy further reduces the classification performances: the accuracy is %78 for (COV=0.3, $\xi$=25), and %69 for (COV=0.5, $\xi$=25).

- The overall performance of ML surrogate models for the classification task on the entire random filed database (all COV and $\xi$ values) when trained on 500 samples (%0.42 of data) is promising with accuracy = %84.7 and AUC = %91.2.

- The errors in the ML predicted probability of failure using 5% of MC data for all heterogeneity and anisotropy levels is below %1 with an average of %0.46 for the entire MC dataset. Such an approach reduced the CPU computational time from 306 days to only 6 hours.

This study portraits the efficiency of ML algorithms as surrogate models in MC simulations when no FOS calculations are required. In another publication (Aminpour, Alaie et al. 2022), we have further explored the efficiency of ML surrogate models in determining the reliability of heterogeneous anisotropic slopes outlining the extent of expected errors.

To improve the performance of ML surrogate models, feature classification (e.g., classifying the input random field data with geometrical aspects within the problem) can be a potential research approach. The dimension reduction techniques applied on the random field data can also be investigated aiming to improve the model performances. Further investigations on the training of deep and convolutional neural networks on random field data with failure, non-failure results (without FOS calculations) can further improve the efficiency of machine learning approaches as a computationally efficient surrogate model for geotechnical reliability analysis.

## Acknowledgement

This research is funded by the Australian Research Council via the Discovery Projects (No. DP200100549).